\pdfoutput=1

\documentclass[11pt]{article}

\usepackage[]{acl}

\usepackage{times}
\usepackage{latexsym}

\usepackage[T1]{fontenc}
\usepackage[ngerman,english]{babel}

\usepackage[utf8]{inputenc}

\usepackage{microtype}

\usepackage{inconsolata}
\usepackage{booktabs}
\usepackage{hyperref}
\usepackage{breakurl}
\addto\extrasenglish{%
}
\title{Cross-Lingual Transfer for Hallucination Detection in German}
\title{HALTEN: Cross-Lingual Transfer for Token-Level \\ Reference-Free Hallucination Detection}
\title{\textsc{anHalten}: Cross-Lingual Transfer for \\ German Token-Level Reference-Free Hallucination Detection}

\author{Janek Herrlein\textsuperscript{1}, Chia-Chien Hung\textsuperscript{2,3}, Goran Glava\v{s}\textsuperscript{1}  \\
\textsuperscript{1}CAIDAS, University of Würzburg, Germany \\
\textsuperscript{2}NEC Laboratories Europe, Heidelberg, Germany\\
\textsuperscript{3}Data and Web Science Group, University of Mannheim, Germany \\
  \texttt{janek.herrlein@live.de}\\
  \texttt{Chia-Chien.Hung@neclab.eu}\\
  \texttt{goran.glavas@uni-wuerzburg.de}\\
  }

\begin{document}
\maketitle
\begin{abstract}
Research on token-level reference-free hallucination detection has predominantly focused on English, primarily due to the scarcity of robust datasets in other languages. This has hindered systematic investigations into the effectiveness of cross-lingual transfer for this important NLP application. To address this gap, we introduce \textsc{anHalten}, a new evaluation dataset that extends the English hallucination detection dataset to German. To the best of our knowledge, this is the first work that explores cross-lingual transfer for token-level reference-free hallucination detection.
~\textsc{anHalten} contains gold annotations in German that are parallel (i.e., directly comparable to the original English instances).
We benchmark several prominent cross-lingual transfer approaches, demonstrating that larger context length leads to better hallucination detection in German, even without succeeding context. Importantly, we show that the sample-efficient few-shot transfer is the most effective approach in most setups. This highlights the practical benefits of minimal annotation effort in the target language for reference-free hallucination detection. Aiming to catalyze future research
on cross-lingual token-level reference-free hallucination detection, we make
\textsc{anHalten} publicly available: \url{https://github.com/janekh24/anhalten}


\end{abstract}

\section{Introduction}

Detecting hallucinations in large pretrained language models~\citep[e.g.,][]{brown2020language, jiang2024mixtral} is critical for ensuring their reliability in real-world applications. Most existing hallucination detection benchmarks focus on \textit{reference-based} tasks (e.g., summarization, machine translation, question answering)~\citep{maynez-etal-2020-faithfulness,rebuffel2022controlling,sadat-etal-2023-delucionqa}, 
comparing model generated text against provided references. However, reference-based hallucination detection is not appropriate for free-form text generation, where obtaining ground-truth references in real-time demands sufficient and accurate preceding retrieval step. To address these challenges, \textit{reference-free} hallucination detection approaches have been introduced~\citep{liu-etal-2022-token,su2024unsupervised}, focusing on identifying inconsistencies within the generated context itself to effectively detect hallucinations in real-time. Besides, most research in hallucination detection has concentrated on \textit{sentence} or \textit{passage}-level~\citep{dhingra-etal-2019-handling, manakul-etal-2023-selfcheckgpt, zhang-etal-2023-sac3}, which is inadequate for real-time applications that require immediate feedback during text generation. Fine-grained, token-level reference-free hallucination detection benchmark is necessary for this purpose. However, research in this area has focused on English~\citep{liu-etal-2022-token}, primarily due to the lack of robust evaluation datasets in other languages. 
Creating token-level hallucination detection datasets for new languages (from scratch or using machine translation) is significantly more expensive and time-consuming than for most other NLP tasks, due to the need for accurate translation and adaptation of nuanced contexts and token-level annotations.
The lack of multilingual evaluation benchmarks hinders the investigation of cross-lingual transfer approaches for token-level reference-free hallucination detection.



\setlength{\tabcolsep}{3.2pt}
\begin{table*}[t]
\centering
\scriptsize{
\begin{tabular}{p{3.76cm} |p{3.76cm} |p{3.76cm} |p{3.76cm}}
\toprule
\multicolumn{1}{c|}{\textbf{\textsc{HaDes}}} & \multicolumn{1}{c|}{\textbf{\textsc{anHalten}}} & \multicolumn{1}{c|}{\textbf{\textsc{HaDes}}} & \multicolumn{1}{c}{\textbf{\textsc{anHalten}}} \\ 
\midrule
\multicolumn{2}{c|}{\textcolor{teal}{\textbf{Not Hallucination}}}& \multicolumn{2}{c}{\textcolor{red}{\textbf{Hallucination}}} \\ \midrule
\tiny
 haunted homes is a british reality television series made by september films productions . [...] the show centers around \textbf{writer richard hillier} ( who owns the rights to the \textbf{story} ) , \textbf{ghostwriter andrew scott smith} ( pilot , only \textbf{aired} due to his lack of confidence \textbf{level} ) [...] 
 they spend \textbf{the weekend} in a supposedly haunted house , \textcolor{teal}{\underline{\textbf{hoping}}} to find out if there are any ghosts around , [...] & \tiny haunted homes ist eine britische reality - fernsehserie , die von september films productions produziert wird . [...] im mittelpunkt der sendung stehen der autor richard hillier  ( der die rechte an der geschichte besitzt ) , der ghostwriter andrew scott smith  ( pilotfilm , der aufgrund seines mangelnden vertrauenslevels nur ausgestrahlt wurde ) [...] 
 sie verbringen das wochenende in einem vermeintlichen spukhaus , \textcolor{teal}{\underline{\textbf{in der hoffnung}}} herauszufinden , ob es dort geister gibt , [...] & \tiny ieva zunda ( born 20 july 1978 in tukums ) is a latvian athlete . 
 [...] she did not make it past the first round at the 1999 and 2003 world championships . [...] in 2008 [...] shortly before the deadline - on 28 june , she had finally reached the \textbf{qualifying} standard in \textbf{the 400 m} ( 56 . 50 ) , as she clocked \textbf{in the first round} . she finished \textbf{third} in her heat , again missing out on a place in the \textcolor{red}{\underline{\textbf{first}}} round . & \tiny ieva zunda ( geboren am 20 . juli 1978 in tukums ) ist eine lettische leichtathletin . 
 [...] bei den weltmeisterschaften 1999 und 2003 kam sie nicht über die erste runde hinaus . [...] 2008 versuchte sie erneut [...] kurz vor dem stichtag - am 28 . juni - hatte sie endlich die qualifikationsnorm über 400 m ( 56 . 50 ) erreicht , wie sie in der ersten runde lief . sie wurde dritte in ihrem lauf und verpasste erneut den einzug in die \textcolor{red}{\underline{\textbf{erste}}} runde . \\ \midrule
 \multicolumn{1}{c|}{Word Spans: [105, 105]}& \multicolumn{1}{c|}{Word Spans: [112, 114]} & \multicolumn{1}{c|}{Word Spans: [153, 153]}& \multicolumn{1}{c}{Word Spans: [154, 154]}\\
\bottomrule
\end{tabular}%
}
\caption{Examples of \textsc{HaDes}~\citep{liu-etal-2022-token} as the perturbed version with token-level label to detect hallucination, and our proposed \textsc{anHalten} machine-translated and post-edited text. The \textbf{bold} terms indicate the perturbed words compared to the original Wiki~\citep{guo-etal-2020-wiki}, and the \underline{underline} term presents the token required to detect hallucination. For brevity, the compared version with original Wiki is available in~\autoref{app:full_data}.}
\label{tab:hades_example}
\end{table*}
\begin{table*}[h]
\centering
\scriptsize{
\begin{tabular}{p{5.11cm} |p{5.12cm} |p{5.11cm}}
\toprule
\multicolumn{1}{c|}{\textbf{ \textsc{HaDes} }} & \multicolumn{1}{c|}{\textbf{\textsc{Machine Translated (MT)}}} & \multicolumn{1}{c}{\textbf{\textsc{anHalten} (MT \& Post-Edited)}} \\ 
\midrule
other similar shows include most haunted and ghost home . it is also shown in the u . s . on the discovery channel fridays and saturdays schedule .
& andere ähnliche shows sind most haunted und ghost home . es ist auch in den \textcolor{blue}{\textbf{u . s .}} auf dem discovery channel freitags und samstags \textcolor{blue}{\textbf{schedule}} gezeigt . & andere ähnliche shows sind most haunted und ghost home . es wird auch in den \textcolor{blue}{\textbf{usa}} auf dem discovery channel freitags und samstags gezeigt . \\ \midrule
dold ' s research in algebraic topology , in particular , his views on fixed - point topology has made him influential in economics as well as mathematics . & dold \textcolor{blue}{\textbf{' s}} forschung in der algebraischen topologie, insbesondere, seine ansichten über fixpunkt-topologie hat ihn einflussreich in der wirtschaft als auch in der mathematik.  & \textcolor{blue}{\textbf{dolds}} forschung in der algebraischen topologie , insbesondere , seine ansichten über fixpunkt - topologie hat ihn \textcolor{blue}{\textbf{sowohl}} in der wirtschaft als auch in der mathematik einflussreich \textcolor{blue}{\textbf{gemacht}} .  \\
\bottomrule
\end{tabular}%
}
\caption{Examples compared with original English \textsc{HaDes} text, the automatic machine translation to German, and the final translation after manual post-editing. The \textcolor{blue}{\textbf{highlighted}} texts indicate the errors that were corrected during post-editing. These errors primarily include incorrect translations, grammatical mistakes, and missing information.}
\label{tab:corrected_sentences}
\end{table*}
In this work, we target this gap and introduce \textbf{\textsc{anHalten}} (germ\textbf{AN} \textbf{HAL}ucina\textbf{T}ion d\textbf{E}tectio\textbf{N}), a new benchmark derived from the English token-level reference-free hallucination detection dataset \textsc{HaDes}~\citep{liu-etal-2022-token}. \textsc{anHalten} is: (1) \textit{reliable} -- with complete texts and hallucination spans (i.e., labels) manually translated, and  (2) \textit{parallel} -- the same set of texts and labels have been translated to German, enabling direct comparison of multilingual models and cross-lingual transfer approaches.

We then use \textsc{anHalten} to benchmark a range of cross-lingual transfer approaches and simulate the real-world applications in multiple setups. Our results show that (i) hallucination detection works comparably well even without succeeding texts, indicating that larger context length helps detect hallucinations in German, thus supporting proactive hallucination prevention \textit{on-the-fly} during text generation, and (ii) few-shot transfer methods achieve high performance with minimal annotated data, highlighting the practical benefits of inexpensive annotation of a handful of target-language hallucination instances for training detection models. 



\section{Methodology}
\subsection{Dataset Creation}
We translate the \textit{full} development set and 10\% of the training set of English \textsc{HaDes} dataset~\citep{liu-etal-2022-token} in German, with 1,000 and 876 instances, respectively.\footnote{Since the original test set labels were not published, we rely on training and development sets throughout our experiments. We also ensure the subsample of the training set retains the original label ratio of the training data.} Each instance includes a \textsc{text}, \textsc{marked word spans}, \textsc{position of marked word spans}, and \textsc{label} to indicate whether the \textsc{marked word spans} causes hallucination. Examples compared to the original English \textsc{HaDes} dataset are shown in~\autoref{tab:hades_example}. 

Following the well-established practice~\citep{hung-etal-2022-multi2woz, senel-etal-2024-kardes}, we carried out a two-phase translation process: (1) we started with an \textit{automatic translation} -- followed by (2) the \textit{manual post-editing} of the translations. We first automatically translate the development and training set portions for both \textsc{text} and \textsc{marked word spans} relying on DeepL Translator. We then incorporate native speaker with University degree and fluent in English, to post-edit the automatic translations to ensure the correctness of the translation -- especially the directly preceding and succeeding context, and the correct determination of the \textsc{marked word spans}. Common errors identified in machine-translated texts include incorrect translations, missing words, grammatical mistakes, or contextual inaccuracies. Examples comparing the original English \textsc{HaDes} with the automatically translated and manually post-edited texts are shown in~\autoref{tab:corrected_sentences}. 
Besides, as the position of \textsc{marked word spans} changes in the German text\footnote{German and English, both Germanic languages, differ in ways that impact dataset design. In German, compound words are written as single words, whereas in English, they are separated by spaces, affecting \textsc{marked word spans}. Additionally, German commonly uses particle verbs, where \textsc{marked word spans} are split by other parts of the sentence. In such cases, only the conjugable main part of the verb is marked, while the particle is ignored.}, the \textsc{position of marked word spans} is adjusted accordingly.

Additionally, to conduct \textit{Translate-Train} experiment for cross-lingual transfer, \textit{full} training set (8,754 instances) are automatically translated using DeepL Translator, without post-editing. However, only 6,344 instances (72.5\%) remain, since the discarded ones contain incorrect \textsc{marked word spans}. 

\subsection{Downstream Cross-Lingual Transfer}
\label{sec:downstream_XLT} The parallel nature and substantial size of \textsc{anHalten} facilitate benchmarking of cross-lingual transfer methods for hallucination detection tasks. We investigate three common methods for downstream cross-lingual transfer (XLT)~\citep{ebing2023translate, senel-etal-2024-kardes}: (1) \textit{Zero-Shot Transfer}, where we assume the absence of labeled task instances in the target language. The model is trained exclusively in English and is expected to perform the task directly in German without prior exposure to German labeled data. This method relies on the model's capability to generalize knowledge from English to German. (2) \textit{Few-Shot Transfer}, where a limited number of labeled instances in the target language exist with the majority of training data in the source language. The model is trained on abundant English data and a small amount of German data jointly,\footnote{Compared to \textit{sequential fine-tuning}~\citep{lauscher-etal-2020-zero, hung-etal-2022-multi2woz}, \textit{joint fine-tuning}~\citep{schmidt-etal-2022-dont} on instances in both source and target language can achieve better performances with higher stability.} helping it adapt to the specific nuances of the German language with limited annotated data. (3) \textit{Translate-Train}, where training instances in source language are automatically translated (i.e., noisy) to target language leveraging the state-of-the-art machine translation model. While this approach relies on the quality of translation, it benefits from creating a substantial amount of training data in German, closely approximating a fully supervised learning scenario. 

To facilitate modular and efficient XLT, adapter-based approach is proposed to learn specialized task and language adapters for high portability and parameter-efficient transfer to various tasks and languages~\citep{pfeiffer-etal-2020-mad}. For downstream XLT, a task adapter is stacked on the pre-trained source language adapter, where the parameters are only updated for the task adapter. During evaluation, the source language adapter is replaced by the pre-trained target language adapter. In our setup, the task adapter is trained by (1) the English-only data for \textit{Zero-Shot Transfer}; (2) a joint training of English and a small portion of German data for \textit{Few-Shot Transfer}; or (3) the machine-translated English-to-German data for \textit{Translate-Train}. The adapter-based approach ensures that the model can efficiently adapt to new tasks with minimal parameter updates, maintaining the balance between performance and computational efficiency. 

\section{Experimental Setup}
\subsection{Evaluation Tasks and Measures}
We evaluate multilingual pre-trained language models (PLMs) in
XLT methods (\autoref{sec:downstream_XLT}) for
token-level reference-free hallucination detection tasks. To simulate real-world applications, we evaluate on two sub-tasks: \textit{offline} and \textit{online}~\citep{liu-etal-2022-token}.  In the \textit{offline} setting, the model accesses both preceding and succeeding contexts of the \textsc{marked word spans}, suitable for detecting hallucinations in pre-generated texts. In the \textit{online} setting, the model considers only the preceding context, enabling proactive prevention of hallucinations during on-the-fly text generation.

We follow~\citet{liu-etal-2022-token} and evaluate the XLT capabilities utilizing multilingual PLMs on hallucination detection tasks. The evaluation metrics include accuracy, precision, recall, F1, Area Under Curve (AUC), G-Mean~\citep{espindola2005extending}, and Brier Score (BS)~\citep{brier1950verification}. These metrics provide a comprehensive evaluation of model performance, balancing correctness, and the ability to handle imbalanced classes.

\subsection{Models and Experimental Setup}
Experiments are conducted on multilingual PLMs, namely multilingual BERT (mBERT)~\citep{devlin-etal-2019-bert} and XLMR~\citep{conneau-etal-2020-unsupervised},\footnote{The weights of PLMs are loaded from HuggingFace: \texttt{multilingual-bert-base-cased} and \texttt{xlm-roberta-base}.} using \textit{language adapters}\footnote{The pre-trained adapters are selected from AdapterHub~\citep{pfeiffer-etal-2020-adapterhub} for English (\texttt{en-wiki@ukp}) and German (\texttt{de-wiki@ukp}).} proposed by~\citet{pfeiffer-etal-2020-mad} to facilitate modular and efficient XLT. 




\setlength{\tabcolsep}{6.2 pt}
\begin{table*}[t]
\centering
\scriptsize{
\begin{tabular}{r c ccccc ccc ccc}
\toprule
& & & & & & & \multicolumn{3}{c}{\textbf{Not Hallucination}} & \multicolumn{3}{c}{\textbf{Hallucination}} \\ \cmidrule(l{.4em}){8-10}\cmidrule(l{.5em}){11-13}
& \textbf{\# Instances}& \textbf{Setting} & \textbf{Accuracy $\uparrow$} & \textbf{G-Mean $\uparrow$} & \textbf{BS $\downarrow$} & \textbf{AUC $\uparrow$} & \textbf{P $\uparrow$} & \textbf{R $\uparrow$} & \textbf{F1 $\uparrow$} & \textbf{P $\uparrow$} & \textbf{R $\uparrow$} & \textbf{F1 $\uparrow$} \\ \midrule
Zero-Shot & 0& offline & 62.82 & 59.38 & 25.51 & 72.90 & \textbf{74.89}& 41.87&53.37 &58.18 &\textbf{84.89} & \textbf{68.96} \\ \midrule
Few-Shot & 10 &offline & 63.86 & 61.89 & 24.23 & 73.32 & 72.84& 47.72&57.34 &59.57 &80.87 &68.51 \\ 
Few-Shot & 100 &offline & 65.12&63.87 &23.28 & 73.88& 72.64& 51.46& 60.12& 60.93&79.51 & 68.94\\ 
Few-Shot  & 876 &offline & \textbf{67.76}&\textbf{67.55} & \textbf{20.83}& \textbf{74.68}&68.24 &69.75 &\textbf{68.87} &67.50 & 65.67&66.44 \\ \midrule 
Translate-Train  & 6344& offline & 66.42 & 64.13 & 21.30 & 73.80 & 66.69& \textbf{72.44}&67.84 &\textbf{69.98} &60.08 &62.91 \\ \midrule
\midrule
Zero-Shot  & 0&online & 63.70&62.01 & 24.14& 72.44&71.51 &48.85 &57.76 &59.72 & \textbf{81.14}&68.02 \\ \midrule
Few-Shot & 10&online & 63.50 & 61.53 & 23.84 & 72.43 & \textbf{72.11}& 47.29&56.88 &59.30 &80.58 &\textbf{68.24} \\ 
Few-Shot& 100&online & 64.88&63.87 &22.88 & 72.55& 70.31& 57.29& 62.03& 61.39&73.93 & 66.19\\
Few-Shot  & 876&online & 67.28&\textbf{67.14} & 21.22& \textbf{73.52}&67.89 &68.89 &68.33 &66.75 & 65.59&66.09 \\ \midrule
Translate-Train &6344& online & \textbf{67.66}&66.55 & \textbf{21.02}& 73.20&65.66 &\textbf{77.66} &\textbf{71.13} &\textbf{70.86} & 57.13&63.19 \\
  \bottomrule
\end{tabular}%
}
\caption{Cross-lingual transfer results of XLMR (\%) averaged over 5 runs. According to \autoref{tab:eval_result_zero_shot_transfer}, XLMR outperforms mBERT. For brevity, cross-lingual transfer results of mBERT are provided in~\autoref{app:additional_experiments}.}
\label{tab:eval_result_few_shot_transfer}
\end{table*}

\setlength{\tabcolsep}{1.0 pt}
\begin{table}[t]
\centering
\tiny{
\begin{tabular}{l ccccc ccc ccc}
\toprule
& & & & & & \multicolumn{3}{c}{\textbf{Not Hallucination}} & \multicolumn{3}{c}{\textbf{Hallucination}} \\ \cmidrule(l{.4em}){7-9}\cmidrule(l{.5em}){10-12}
\textbf{Model} & \textbf{Setting} & \textbf{Acc. $\uparrow$} & \textbf{G-Mean $\uparrow$} & \textbf{BS $\downarrow$} & \textbf{AUC $\uparrow$} & \textbf{P $\uparrow$} & \textbf{R $\uparrow$} & \textbf{F1 $\uparrow$} & \textbf{P $\uparrow$} & \textbf{R $\uparrow$} & \textbf{F1 $\uparrow$} \\ \midrule
mBERT & offline & 61.00 & 56.12 & 26.49 & 69.66 & 71.04 & 42.92 & 50.68 & 57.89 & 80.04 & 66.54 \\
XLMR & offline & 62.82 & 59.38 & 25.51 & \textbf{72.90} & \textbf{74.89}& 41.87&53.37 &58.18 &84.89 &\textbf{68.96} \\ 
\midrule 
mBERT & online & 60.44&55.34 &26.71 & 67.81& 73.47& 36.69& 48.10& 56.33&\textbf{85.46} & 67.76\\
XLMR & online & \textbf{63.70}&\textbf{62.01} & \textbf{24.14}& 72.44&71.51 &\textbf{48.85} &\textbf{57.76} &\textbf{59.72} & 81.14&68.02 \\
  \bottomrule
\end{tabular}%
}
\caption{Zero-shot transfer results (\%) averaged over 5 runs. Reference English performance of XLMR for accuracy: 70.40\% (\textit{offline}), 68.80\% (\textit{online}).}
\label{tab:eval_result_zero_shot_transfer}
\end{table}
To evaluate downstream XLT, the experiments are conducted with 5 runs in both \textit{offline} and \textit{online} settings, with a fixed context window of 200 tokens. In the \textit{online} setting, the context includes the 200 tokens preceding the \textsc{marked word spans}. In the \textit{offline} setting, it includes 100 tokens before and after the \textsc{marked word spans}.\footnote{\citet{liu-etal-2022-token} observed that model performance for English \textsc{HaDes} dataset stabilizes around 80 tokens, with minimal performance differences between \textit{offline} and \textit{online} settings regarding context length. Thus, using 200 tokens would not limit performance, and increasing the context is unlikely to improve results.} During training, the instances are randomly split into a 70/30 train and validation split, while the original label ratio of training data is retained for the split. We train for 10 epochs in batches of 8 instances, with learning rate $5\cdot10^{-3}$, and a dropout ratio 0.2 is set to avoid overfitting. 
\section{Results and Discussions}
We present and discuss the downstream XLT results on \textsc{anHalten} for the token-level reference-free hallucination detection task across three XLT setups (\autoref{sec:downstream_XLT}): zero-shot transfer, few-shot transfer, and translate-train.

\paragraph{Zero-Shot Transfer.} The results summarized in \autoref{tab:eval_result_zero_shot_transfer} highlight the performance of zero-shot transfer. Notably, XLMR consistently outperforms mBERT across most metrics, indicating that XLMR is better suited for zero-shot transfer. Minimal performance differences between the online and offline settings suggest that the selection of large context windows does not significantly impact performance, aligning with findings from~\citet{liu-etal-2022-token}. Having only preceding text with larger context lengths aids in detecting hallucinations, which is valuable for real-world applications, especially for proactively preventing hallucinations during on-the-fly generation. Compared with reference English performance, the zero-shot transfer results show significantly lower accuracy for both online and offline settings, with drops exceeding 5\% points. These substantial performance declines underscore the inherent challenges in achieving reliable zero-shot XLT, which is consistent with the findings from prior work~\citep{lauscher-etal-2020-zero, pfeiffer-etal-2020-mad}.
\paragraph{Few-Shot Transfer and Translate-Train.}
As detailed in~\autoref{tab:eval_result_few_shot_transfer}, few-shot transfer results for XLMR show remarkable improvements as the number of annotated German instances increases. With 10\% of the English \textsc{HaDEs} training set (i.e., 876 annotated instances), accuracy improves by 4.9\% points (offline) and 3.6\% points (online) compared to zero-shot transfer. The corresponding G-Mean score increases by 8.2\% points (offline) and 5.1\% points (online). Notably, with only 100 annotated instances, accuracy improves by 2.3\% points (offline) and 1.2\% points (online), and the G-Mean score improves by 4.5\% points (offline) and 1.9\% points (online). This demonstrates the substantial impact of incorporating minimal annotated data on enhancing XLT performance. The translate-train approach, which involves translating a large corpus of 6,344 instances, yields accuracy gains of 3.6\% points (offline) and 4.0\% points (online) compared to zero-shot transfer. While beneficial, the marginal gains compared to few-shot transfer highlight the practical efficiency of using smaller amounts of high-quality annotated data. Based on our findings, few-shot transfer emerges as a highly viable strategy for cross-lingual transfer of reference-free hallucination detection, offering robust performance gains over zero-shot transfer without the extensive resource required by the translate-train approach. This re-emphasizes the well-documented practical benefits of few-shot cross-lingual transfer \cite{lauscher-etal-2020-zero,schmidt-etal-2022-dont}, here for reference-free hallucination detection. 

\setlength{\tabcolsep}{1.1 pt}
\begin{table}[t]
\centering
\tiny{
\begin{tabular}{l cccc ccc ccc}
\toprule
& & & & & \multicolumn{3}{c}{\textbf{Not Hallucination}} & \multicolumn{3}{c}{\textbf{Hallucination}} \\ \cmidrule(l{.4em}){6-8}\cmidrule(l{.5em}){9-11}
\textbf{POS} & \textbf{Accuracy $\uparrow$} & \textbf{G-Mean $\uparrow$} & \textbf{BS $\downarrow$} & \textbf{AUC $\uparrow$} & \textbf{P $\uparrow$} & \textbf{R $\uparrow$} & \textbf{F1 $\uparrow$} & \textbf{P $\uparrow$} & \textbf{R $\uparrow$} & \textbf{F1 $\uparrow$} \\ \midrule
Adjectives  & \textbf{65.80} & \textbf{64.65} & \textbf{22.20} & \textbf{72.61} & \textbf{71.07} & 53.55 & 61.06 & \textbf{62.66} & \textbf{78.06} & \textbf{69.52} \\
Nouns  & 58.42 & 56.31 & 25.37 & 63.99 & 61.97& 43.97&51.15 &56.64 &72.88 &63.62 \\
Verbs  & 52.16&37.20 &28.13 & 59.69& 51.24& \textbf{88.65}& \textbf{64.93}& 58.78&15.68 & 24.64\\
  \bottomrule
\end{tabular}%
}
\caption{Part-of-Speech (POS) results of XLMR (\%) in the \textit{online} setting averaged over 5 runs. We only consider instances with \textsc{marked word spans} containing particular types of POS in the German language: adjectives, nouns, verbs.}
\label{tab:eval_result_pos_german}
\end{table}
\paragraph{Analysis.} According to~\citet{liu-etal-2022-token}, nouns and verbs are the most frequently occurring part-of-speech (POS) in the \textsc{marked word spans} of the HADES dataset. The majority of instances with nouns (62.4\%) and adjectives (74.0\%) in the \textsc{marked word spans} belong to the hallucination class, while the majority of instances with verbs belong to the non-hallucination class (62.8\%). This indicates a significant imbalance in label distribution. To assess the impact of this imbalance on cross-lingual transfer performance, we classify the validation set of \textsc{anHalten} based on the selected POS (nouns, verbs, adjectives) in the \textsc{marked word spans}. Instances with \textsc{marked word spans} containing multiple words from different POS are excluded. To ensure an equal number of labels for each POS, we randomly remove instances from the more frequent class. This process results in 292 noun instances, 222 verb instances, and 62 adjective instances.


We then analyze the XLT results of XLMR in the online setting. The POS results in \autoref{tab:eval_result_pos_german} show that adjectives are significantly more effective in detecting hallucinations compared to nouns and verbs in German. While the effectiveness of adjectives is notable, the imbalanced distribution of instances across different part-of-speech tags, as highlighted by \citet{liu-etal-2022-token}, warrants further investigation and consideration. Addressing these imbalances is crucial for improving the overall robustness and accuracy of hallucination detection models.

We further conduct morphological analysis (detailed in~\autoref{app:additional_experiments}) and demonstrate that preceding words indicate grammatical gender in German impact model performance, underscoring the importance of linguistic context. These findings emphasize the need to address imbalances and encourage future work to enhance model performance concerning diverse linguistic features for token-level reference-free hallucination detection.

\section{Conclusions}
Token-level reference-free hallucination detection has predominantly focused on English, primarily due to the lack of robust benchmarks in other languages, hindering investigation into cross-lingual transfer approaches for this important task. To address this gap, we have presented~\textsc{anHalten}, an extension of the English \textsc{HaDes} containing gold hallucination annotations in German, allowing for reliable and comparable cross-lingual estimates for token-level reference-free hallucination detection tasks. We utilized a modular adapter-based approach to facilitate the cross-lingual transfer, demonstrating the effectiveness of sample-efficient few-shot transfer. We believe that our dataset and findings advance the understanding of hallucination detection in cross-lingual transfer setups and contribute towards multilingual hallucination detection and real-time hallucination prevention in free-form text generation.
\section*{Acknowledgements}
This work was supported by the Alcatel-Lucent Stiftung and Deutsches Stiftungszentrum through the grant ``Equitably Fair and Trustworthy Language Technology'' (EQUIFAIR, Grant Nr. T0067/43110/23).  

\section*{Limitations}
Despite the contributions of this research, several limitations are acknowledged, which present opportunities for future enhancement. Currently, \textsc{anHalten} extends hallucination detection to German, broadening the scope beyond English but still covering only two languages. Expanding this research to include additional languages could further increase the global applicability of our findings. Besides, incorporating data from sources other than Wikipedia could enrich the diversity and complexity of the dataset. Additionally, extending the research to include other types of hallucinations (e.g., subjective hallucinations) would provide a more comprehensive understanding of hallucination detection in various text types. We experimented on encoder-only multilingual PLMs, while decoder-based PLMs~\citep[e.g.,][]{scao-etal-2022-bloom, jiang2023mistral, abdin2024phi} warrants exploration. We hope that future research builds on top of our findings and
extends the research toward more domains, more languages, and specifically with the
efficiency and effective concerns of hallucination detection in different languages. 
\section*{Ethics Statement}
This research addresses the critical need for non-English language datasets in hallucination detection by introducing \textsc{anHalten}. The ethical considerations of this work are multifaceted. By extending hallucination detection to German, the research promotes linguistic diversity and inclusivity in AI systems. This inclusivity helps to mitigate biases and misinformation that can arise from language restrictions, fostering more equitable applications. The study also aims to facilitate the recognition of potential hallucinated content produced by large-scale pretrained models in free-form generation -- could be useful in both \textit{offline} and \textit{online} settings. Additionally, the research outcome emphasizes the importance of resource-efficient approaches, reducing the reliance on extensive annotated data and promoting more sustainable development.
\bibliography{custom}

\appendix
\clearpage
\onecolumn
\label{sec:appendix}
\section{Dataset}
The \textsc{HaDes} dataset, introduced by~\citet{liu-etal-2022-token}, is designed for reference-free token-level hallucination detection tasks in English. It is sourced from English Wikipedia~\citep{guo-etal-2020-wiki}, with extracted text segments that are first perturbed and then verified by crowd-sourced annotators to determine if the marked word spans in the text cause hallucination. The dataset is available under an open-source MIT License and contains a total of 10954 instances, divided into train, development, and test sets with sizes of 8754, 1000, 1200 respectively. Within the dataset, 54.5\% of the instances are classified as \textit{Hallucination}, while 45.5\% of the instances are classified as \textit{Not Hallucination}. Since the original labels of the test set were not published, we primarily rely on the training and development sets throughout our experiments. 

To further facilitate research on cross-lingual transfer in German hallucination detection tasks, we propose \textsc{anHalten} in this work. We manually annotate 876 training instances and the entire development set of 1000 instances, which are machine-translated and further post-edited. The proposed \textsc{anHalten} compared with the original Wiki and perturbed \textsc{HaDES}, is shown in~\autoref{tab:hades_example_full}.

\label{app:full_data}
\setlength{\tabcolsep}{4.8pt}
\begin{table*}[h]
\centering
\scriptsize{
\begin{tabular}{p{5.0cm} |p{5.0cm} |p{5.0cm}}
\toprule
\multicolumn{1}{c|}{\textbf{ Wiki (Original) }} & \multicolumn{1}{c|}{\textbf{\textsc{HaDes} (Perturbed)}} & \multicolumn{1}{c}{\textbf{\textsc{anHalten} (MT \& Post-Edited)}} \\ 
\midrule
& \multicolumn{2}{c}{\textcolor{teal}{\textbf{Not Hallucination}}}  \\ \midrule
haunted homes is a british reality television series made by september films productions . [...] the show centers around \textbf{psychic mia dolan} ( who owns the rights to the \textbf{programme} ) , \textbf{ghost hunter david vee} ( pilot \textbf{episode} , only \textbf{allegedly} due to his lack of confidence \textbf{presenting} ) , actor \textbf{mark webb} and \textbf{professor / sceptic} chris \textbf{french} . they spend \textbf{two nights} in a supposedly haunted house , hoping to find out if there are any ghosts around , [...]
& haunted homes is a british reality television series made by september films productions . [...] the show centers around \textbf{writer richard hillier} ( who owns the rights to the \textbf{story} ) , \textbf{ghostwriter andrew scott smith} ( pilot , only \textbf{aired} due to his lack of confidence \textbf{level} ) , actor \textbf{paul newman} and \textbf{scientist / paranormal investigation officer} chris \textbf{martin} . they spend \textbf{the weekend} in a supposedly haunted house , \textcolor{teal}{\underline{\textbf{hoping}}} to find out if there are any ghosts around , [...] & haunted homes ist eine britische reality - fernsehserie , die von september films productions produziert wird . [...] im mittelpunkt der sendung stehen der autor richard hillier  ( der die rechte an der geschichte besitzt ) , der ghostwriter andrew scott smith  ( pilotfilm , der aufgrund seines mangelnden vertrauenslevels nur ausgestrahlt wurde ) , der schauspieler paul newman und der wissenschaftler / paranormale untersuchungsbeauftragte chris martin . sie verbringen das wochenende in einem vermeintlichen spukhaus , \textcolor{teal}{\underline{\textbf{in der hoffnung}}} herauszufinden , ob es dort geister gibt , [...] \\ \midrule
& Word Spans: [105, 105] & Word Spans: [112, 114] \\ \midrule \midrule
& \multicolumn{2}{c}{\textcolor{red}{\textbf{Hallucination}}} \\\midrule
ieva zunda ( born 20 july 1978 in tukums ) is a latvian athlete . her main event is the sprint and hurdles , but she also competes in the 400 and 800 metres . [...] she did not make it past the first round at the 1999 and 2003 world championships . [...] in 2008 [...] shortly before the deadline - on 28 june , she had finally reached the \textbf{entry} standard in \textbf{400 m hurdles} ( 56 . 50 ) , as she clocked \textbf{56.34 seconds} , she finished \textbf{fifth} in her heat , again missing out on a place in the \textcolor{red}{\underline{\textbf{second}}} round . & ieva zunda ( born 20 july 1978 in tukums ) is a latvian athlete . her main event is the sprint and hurdles , but she also competes in the 400 and 800 metres . [...] she did not make it past the first round at the 1999 and 2003 world championships . [...] in 2008 [...] shortly before the deadline - on 28 june , she had finally reached the \textbf{qualifying} standard in \textbf{the 400 m} ( 56 . 50 ) , as she clocked \textbf{in the first round} . she finished \textbf{third} in her heat , again missing out on a place in the \textcolor{red}{\underline{\textbf{first}}} round . & ieva zunda ( geboren am 20 . juli 1978 in tukums ) ist eine lettische leichtathletin . ihre hauptdisziplin ist der sprint und der hürdenlauf , sie tritt aber auch über 400 und 800 m an . [...] bei den weltmeisterschaften 1999 und 2003 kam sie nicht über die erste runde hinaus . [...] 2008 versuchte sie erneut [...] kurz vor dem stichtag - am 28 . juni - hatte sie endlich die qualifikationsnorm über 400 m ( 56 . 50 ) erreicht , wie sie in der ersten runde lief . sie wurde dritte in ihrem lauf und verpasste erneut den einzug in die \textcolor{red}{\underline{\textbf{erste}}} runde . \\ \midrule
& Word Spans: [153, 153] & Word Spans: [154, 154]  \\
\bottomrule
\end{tabular}%
}
\caption{Examples of the original text from Wikipedia~\citep{guo-etal-2020-wiki}, \textsc{HaDes}~\citep{liu-etal-2022-token} as the perturbed version with token-level labels to detect hallucination, and the machine-translated (MT) and post-edited version from our proposed \textsc{anHalten}.}
\label{tab:hades_example_full}
\end{table*}
\newpage
\section{Additional Experiments}
\label{app:additional_experiments}
\subsection{Cross-Lingual Transfer Results of mBERT}
\setlength{\tabcolsep}{3.1 pt}
\begin{table*}[h]
\centering
\footnotesize{
\begin{tabular}{r c ccccc ccc ccc}
\toprule
 & & & & & & & \multicolumn{3}{c}{\textbf{Not Hallucination}} & \multicolumn{3}{c}{\textbf{Hallucination}} \\ \cmidrule(l{.4em}){8-10}\cmidrule(l{.5em}){11-13}
& \textbf{\# Instances}& \textbf{Setting} & \textbf{Accuracy $\uparrow$} & \textbf{G-Mean $\uparrow$} & \textbf{BS $\downarrow$} & \textbf{AUC $\uparrow$} & \textbf{P $\uparrow$} & \textbf{R $\uparrow$} & \textbf{F1 $\uparrow$} & \textbf{P $\uparrow$} & \textbf{R $\uparrow$} & \textbf{F1 $\uparrow$} \\ \midrule
Zero-Shot & 0& offline & 61.00 & 56.12 & 26.49 & 69.66 & 71.04 & 42.92 & 50.68 & 57.89 & 80.04 & 66.54 \\ \midrule
Few-Shot &10 & offline & 62.84 & 60.36 & 25.96 & 69.53 & 71.41& 46.94&55.74 &59.15 &79.59 &\textbf{67.54} \\ 
Few-Shot &100 & offline & 61.16&55.94 &25.52 & 70.27& \textbf{73.14}& 40.98& 49.95& 58.02&\textbf{82.42} & 67.21\\ 
Few-Shot  & 876& offline & 62.08&58.61 & 24.27& 70.44&68.46 &52.63 &56.48 &60.80 & 72.03&64.29 \\ \midrule 
Translate-Train &6344& offline & \textbf{64.54} & \textbf{63.76} & \textbf{22.61} & \textbf{70.46} & 66.97 & \textbf{62.61} & \textbf{63.99} & \textbf{63.44} & 66.57 & 64.35 \\ \midrule 
\midrule
Zero-Shot  &0& online & 60.44&55.34 &26.71 & 67.81& \textbf{73.47}& 36.69& 48.10& 56.33&\textbf{85.46} & \textbf{67.76} \\ \midrule
Few-Shot &10& online & 60.04 & 53.77 & 27.61& 68.05 & 72.94& 36.53&46.48 &56.46 &84.81 &67.39 \\ 
Few-Shot &100& online & 60.84&56.55 &27.35 & 67.44& 70.99& 41.79& 50.68& 57.32&80.90 & 66.77\\
Few-Shot  &876& online & 61.50&58.07 & 24.66& 68.86&71.52 &43.66 &52.82 &57.91 & 80.29&66.82 \\ \midrule
Translate-Train &6344& online & \textbf{64.78}&\textbf{63.66} &\textbf{22.66} & \textbf{71.11}& 69.43& \textbf{57.66}& \textbf{62.05}& \textbf{62.55}&72.28 & 66.45\\
  \bottomrule
\end{tabular}%
}
\caption{Cross-lingual transfer results of mBERT (\%) averaged over 5 runs.}
\label{tab:eval_result_few_shot_transfer_mbert}
\end{table*}
\subsection{Morphological Analysis}
In English, grammatical gender is not distinguished, whereas German has three grammatical genders that influence articles, pronouns, and adjectives. Words indicating gender often lie outside the \textsc{marked word spans} used for hallucination detection. Our experiment selects instances where gender-indicating words (articles, possessive pronouns, demonstrative pronouns) precede nouns in the \textsc{marked word spans} from both the German and English datasets. This dataset includes 64 instances per language, with an equal distribution of labels.

Testing with XLMR in the online setting, the goal is to determine if contextual gender information influences hallucination detection results. The additional gender information might help classify non-hallucination instances but could mislead models if the original, correct word has a different gender.

Results in~\autoref{tab:eval_result_morphological} show a performance drop in accuracy and G-Mean when gender-indicating words are included in the \textsc{marked word spans}, particularly for English instances. However, AUC improves, suggesting that the extended spans do not hinder the model's ability to distinguish between classes. The models tend to assign more instances to the hallucination class, reducing the F1 score for the non-hallucination class. This performance drop may result from a lack of such gender-indicating contexts in the fine-tuning dataset, indicating potential issues with handling longer \textsc{marked word spans}.

\setlength{\tabcolsep}{5.3 pt}
\begin{table*}[h]
\centering
\footnotesize{
\begin{tabular}{l ccccc ccc ccc}
\toprule
& & & & & & \multicolumn{3}{c}{\textbf{Not Hallucination}} & \multicolumn{3}{c}{\textbf{Hallucination}} \\ \cmidrule(l{.4em}){7-9}\cmidrule(l{.5em}){10-12}
\textbf{Language} & \textbf{Preceding} & \textbf{Accuracy $\uparrow$} & \textbf{G-Mean $\uparrow$} & \textbf{BS $\downarrow$} & \textbf{AUC $\uparrow$} & \textbf{P $\uparrow$} & \textbf{R $\uparrow$} & \textbf{F1 $\uparrow$} & \textbf{P $\uparrow$} & \textbf{R $\uparrow$} & \textbf{F1 $\uparrow$} \\ \midrule
English & With & 76.56 & 76.13 & 16.74 & 85.16 & 76.73 & 76.88 & 76.31 & 77.94 & 76.25 & 76.51 \\
German & With & 59.38 & 46.32 & 21.73 & 86.99 & 85.66& 24.37&35.99 &55.75 &94.37 &69.93 \\ \midrule
English & Without & 69.69&68.58 &21.67 & 74.57& 75.30& 59.38& 65.97& 66.67&80.00 &72.46\\
German & Without & 61.25&49.38 & 22.34& 80.96&86.59 &27.50 &39.63 &57.17 & 95.00&71.16 \\
  \bottomrule
\end{tabular}%
}
\caption{Results of XLMR (\%) in the \textit{online} setting averaged over 5 runs, for instances with \textsc{marked word spans} containing nouns with and without preceding words that indicate the grammatical gender of the noun.}
\label{tab:eval_result_morphological}
\end{table*}

\end{document}